# MULTILINGUAL BERT LANGUAGE MODEL FOR MEDICAL TASKS: EVALUATION ON DOMAIN-SPECIFIC ADAPTATION AND CROSS-LINGUALITY

Yinghao Luo[1,2], Lang Zhou[1,2], Amrish Jhingoer[1,2],
Klaske Vliegenthart–Jongbloed[3,4], Carlijn Jordans[4],
Ben Werkhoven[5], Tom Seinen[6], Erik van Mulligen[6],
Casper Rokx[3,4], and Yunlei Li[1*]

[1]Department of Pathology & Clinical Bioinformatics, Erasmus University Medical Center Rotterdam
[2]Department of Computer Science, Vrije Universiteit Amsterdam
[3]Department of Internal Medicine, Erasmus University Medical Center Rotterdam
[4]Department of Medical Microbiology & Infectious Diseases, Erasmus University Medical Center Rotterdam
[5]Department of Data & Analytics, Erasmus University Medical Center Rotterdam
[6]Department of Medical Informatics, Erasmus University Medical Center Rotterdam

## ABSTRACT

In multilingual healthcare applications, the availability of domain-specific natural language processing(NLP) tools is limited, especially for low-resource languages. Although multilingual bidirectional encoder representations from transformers (BERT) offers a promising motivation to mitigate the language gap, the medical NLP tasks in low-resource languages are still underexplored. Therefore, this study investigates how further pre-training on domain-specific corpora affects model performance on medical tasks, focusing on three languages: Dutch, Romanian and Spanish. In terms of further pre-training, we conducted four experiments to create medical domain models. Then, these models were fine-tuned on three downstream tasks: Automated patient screening in Dutch clinical notes, named entity recognition in Romanian and Spanish clinical notes. Results show that domain adaptation significantly enhanced task performance. Furthermore, further differentiation of domains, e.g. clinical and general biomedical domains, resulted in diverse performances. The clinical domain-adapted model outperformed the more general biomedical domain-adapted model. Moreover, we observed evidence of cross-lingual transferability. Moreover, we also conducted further investigations to explore potential reasons contributing to these performance differences. These findings highlight the feasibility of domain adaptation and cross-lingual ability in medical NLP. Within the low-resource language settings, these findings can provide meaningful guidance for developing multilingual medical NLP systems to mitigate the lack of training data and thereby improve the model performance.

## 1 Introduction

### 1.1 Motivation

Human immunodeficiency virus (HIV) remains a major challenge to public health on a global scale. Currently, although universal access to antiretroviral therapy has significantly reduced the mortality and transmission risk caused by HIV, many individuals still suffer health risks due to late diagnosis or a lack of timely treatment. Late HIV diagnosis is associated with adverse outcomes, including increased morbidity, mortality, and onward risk of transmission[1]. These negative consequences affect not only the individual but also the broader population [2]. Hence, it is essential to minimize missed opportunities for early HIV screening.

---
[*]y.li.1@erasmusmc.nl

With the rapid expansion of electronic health record (EHR) applications, there is great potential to use artificial intelligence tools to automatically identify patients who might be at risk or need testing. Traditionally, extracting useful information from unstructured data relies on manual chart review by experts, which would be extremely time-consuming and labor-intensive [3]. In recent years, natural language (NLP) techniques, particularly pre-trained language models such as BERT [4] and its domain-specific adaptations, have been widely applied to multiple healthcare tasks, such as disease information extraction, risk prediction, and treatment recommendations [5].

Many healthcare settings involve multilingual contexts or under-resourced languages, such as Romanian and Spanish. It is worth mentioning that Romanian and Spanish both face significant challenges due to the scarcity of medical NLP resources [6, 7]. Multilingual language models are designed to understand and process multiple languages by sharing a common vocabulary across diverse linguistic unlabeled data. This property makes them particularly useful in cross-lingual transfer settings, especially applying to low-resource languages [8]. By combining language models with medical knowledge with the ability to work on different languages, they show potential to improve performance on medical tasks, in regions where relevant data is limited.

However, despite these advancements, existing research for medical tasks has mostly focused on English-language medical data, with limited attention to under-resourced languages. Furthermore, the potential of pre-trained multilingual BERT-based models for supporting HIV screening in real-world EHR data remains underexplored. Moreover, the effectiveness of cross-lingual transfer for low-resource languages, using models trained on more abundant clinical corpus (such as in Dutch), is also underexplored. Addressing these gaps is crucial for extending the benefits of automated tools to more diverse healthcare settings.

## 1.2 Problem statement

Although the use of pre-trained language models has advanced benefits for medical tasks, several important challenges remain unresolved.

First of all, there is limited understanding of how further pre-training on in-domain corpora affects the performance of mBERT on downstream medical NLP tasks. While domain specific models such as BioBERT [9], ClinicalBERT [10] have shown great progress in English, their performance remains poor in most low-resource languages. Therefore, we chose mBERT as our baseline model. Afterwards, the relative effectiveness of different types of domain-specific datasets (e.g., biomedical vs. clinical corpora) in improving model performance is still underexplored. Ultimately, the potential of cross-lingual transfer, which involves leveraging models pre-trained on more abundant clinical corpora (such as in Dutch) to support downstream tasks in low-resourced languages (Romanian and Spanish), remains an open research question.

To address the gaps identified in previous research, this study proposes the following hypotheses. First, domain adaptation is expected to improve the performance of mBERT on medical NLP tasks. Second, further pre-training on clinical Dutch corpora is hypothesized to yield better performance on medical NLP tasks in Dutch compared to pre-training on general biomedical Dutch corpora. Third, pre-training on Dutch as a source language is anticipated to enhance performance on downstream tasks in other target languages, indicating cross-lingual transferability.

## 1.3 Contributions

This study developed a Dutch clinical language model and evaluated it on the HIV screening task. To our knowledge, this is the first Dutch clinical language model further pre-trained for HIV screening task. In addition, we systematically investigated how further pre-training on in-domain corpora affects model performance on downstream medical tasks. In particular, we also investigated how further pre-training on either general biomedical or clinical Dutch data influences model performance on the HIV classification task. Furthermore, we analyzed the cross-lingual ability of model that was pre-trained on Dutch clinical corpora and subsequently applied to Romanian and Spanish downstream tasks, providing insights into the potential and limitations of multilingual BERT-based models for cross-lingual ability. Through comprehensive empirical experiments and comparative analysis, this work contributed to a better understanding of multilingual language models in diverse and under-resourced healthcare tasks.

## 1.4 Outline

The remainder of this article is organized as follows. In section 2, we demonstrate the background and related work of language models, domain adaptation and cross-lingual transfer in the medical domain. Section 3 describes the materials

Multilingual BERT Language Model for Medical Tasks: Evaluation on Domain-Specific Adaptation and Cross-linguality

and methods in every single experiment, which also includes the assumptions we propose. Then, section 4 gives a detailed analysis of the experimental results. Additionally, section 5 analyzes the findings with the comprehensive justification. Finally, section 6 concludes the findings, our limitations and future plan in this study.

## 2 Related Work

### 2.1 Transformer Architecture

The transformer, proposed by Vaswani et al.[11], marks a major shift in natural language processing by replacing recurrent neural networks with a fully attention-based mechanism. It relies entirely on self-attention mechanisms to model dependencies between tokens, allowing for significantly improved parallelization and performance on sequence modeling tasks.

As shown in Figure 1, the left side is the encoder, while the right is the decoder. The encoder structure is composed of a stack of six identical encoder layers, each layer comprises a multi-head attention sub-layer and the position-wise fully connected feed forward sub-layer. These two components are wrapped with residual connections and layer normalization, enabling better gradient flow and training stability. The difference of the decoder structure with encoder structure is the additional multi-head attention sub-layer. The additional multi-head sub-layer is connected to the output of each encoder, which allows the decoder to attend the encoder's output representations. The encoder processes the input sequence and generates contextualized token representations. The decoder uses both previously generated tokens via self-attention and the encoder output via cross-attention to predict the next token in the output sequence, enabling effective sequence-to-sequence modeling.

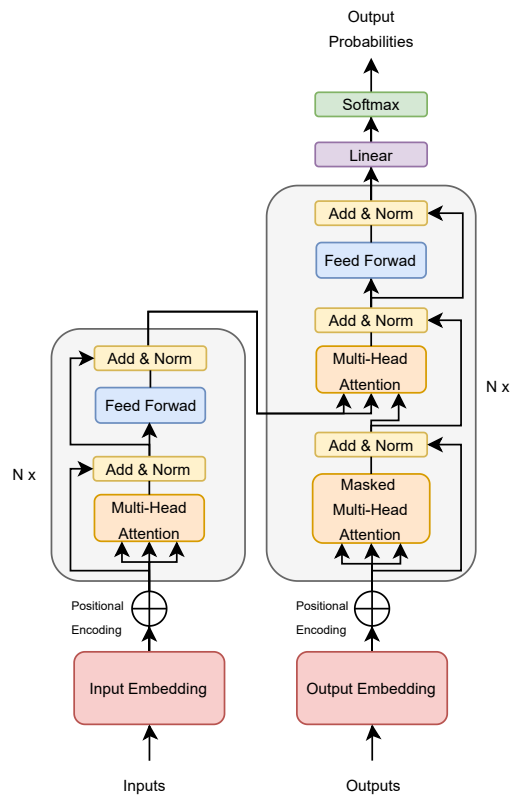

Figure 1: The architecture of the transformer. [4]

### 2.1.1 Self-Attention Mechanism

The self-attention mechanism of the transformer architecture is referred to as the scaled dot-product attention by Vaswani et al. [11]. The mechanism begins by projecting the input sequence into three matrices: queries (Q), keys (K) and values (V). These are obtained by multiplying the input embeddings with learned weight matrices. Conceptually, the query represents the token we are focusing on, the keys represent the tokens it is compared to, and the values contain the information that will be aggregated.

The similarity between a query and each key is measured using a dot product, which captures how much attention one token should pay to another. These similarity scores are then scaled by the inverse square root of the key dimension to prevent the values from growing too large, which can lead to vanishing gradients during training. After scaling, a softmax function is applied to convert the similarity scores into a probability distribution. These probabilities serve as attention weights, indicating how much each token contributes to the representation of the current token. The final output is computed as a weighted sum of the value vectors, where the weights are the attention scores. Below is the mathematical expression:

$$Attention(Q, K, V) = softmax(\frac{QK^T}{\sqrt{d_k}})V \quad (1)$$

This mechanism enables each position in the sequence to attend selectively to other positions, regardless of their distance, making it possible to model long-range dependencies more efficiently than with recurrent architectures. Furthermore, it is fully parallelizable, which greatly improves computational efficiency.

### 2.1.2 Multi-Head attention Mechanism

While a single attention layer can help models attend to relevant tokens based on contextual similarity, relying on a single space might limit the expressive power of models. To better train the model, the multi-head attention mechanism has been applied to the transformer, which allows transformer to jointly attend to information from different representation subspaces at different positions. Initially, Q, K and V are linearly projected h times to $d_Q$, $d_K$ and $d_V$ respectively. Then, queries, keys and values on each dimension will be input into the scaled dot-product attention layer in parallel. Finally, these outputs on h dimensions are concatenated and projected once more through a final linear layer.

$$head_i = Attention(QW_i^Q, KW_i^K, VW_i^V) \quad (2)$$

$$MultiHead(Q, K, V) = Concat(head_1, ..., head_h)W^O \quad (3)$$

Where the projection matrices are defined as below:

$$W_i^Q, W_i^K \in \mathbb{R}^{d_{model} \times d_k}, \quad W_i^V \in \mathbb{R}^{d_{model} \times d_v}, \quad W^O \in \mathbb{R}^{hd_v \times d_{model}}$$

This mechanism allows the model to aggregate information from multiple dimensions simultaneously, greatly enhancing its capacity to represent complex linguistic patterns. By distributing attention across multiple heads, the Transformer can encode both local and global dependencies more effectively than a single-head architecture.

## 2.2 BERT

Bidirectional Encoder Representation transformer (BERT) is a pre-trained language model introduced by Devlin et al. [4], which has brought a wide range of state-of-the-art results in NLP tasks. The architecture of BERT consists of multiple layers of bidirectional self-attention and feed-forward networks, allowing it to capture contextual information from both the left and right of each token. BERT is pre-trained on large-scale unlabeled text using two unsupervised objectives: masked language models (MLM) and next sentence prediction (NSP).

The concept of masked language modeling is inspired by the Cloze task, originally proposed by Taylor et al. [12], in which words are removed from a text and must be predicted based on context. As implemented in BERT, 15% of the input tokens from WordPiece are randomly selected for prediction. Of those selected tokens, 80% tokens are replaced with the special [MASK] token, 10% are substituted with a random token from the vocabulary, and the remaining 10% are left unchanged. After this, the model is trained to predict the original identity of these masked tokens using bidirectional context, while ensuring that the model does not have access to the true identity of the masked token. This formulation allows the model to learn contextualized token representations based on both left and right context, without relying on sequential prediction.

Apart from masked language modeling, BERT introduces the other pre-training objective known as Next Sentence Prediction (NSP), which is designed to model inter-sentential relationships. This task is particularly relevant for Multilingual BERT Language Model for Medical Tasks: Evaluation on Domain-Specific Adaptation and Cross-linguality downstream tasks such as question answering and natural language inference, where understanding the relationship between sentences is of importance. Specifically, 50% of the sentence pairs consist of a consecutive sentence B that follows sentence A in the original corpus (labeled as IsNext), while the remaining 50% are randomly paired sentences that do not occur adjacent to one another (labeled as NotNext). The model is trained to predict if sentence B is the next sentence of sentence A using a classification layer, which enables the model to capture higher-level coherence between sentence pairs during pre-training.

Although NSP is introduced in BERT to improve model's ability to understand inter-sentential relationships, subsequent research has doubted its necessity [13, 14]. Liu et al. use MLM and remove NSP entirely, but the result still achieves superior performance on multiple downstream NLP tasks [14]. Most researchers apply MLM method to further pre-train their own BERT models [15, 13, 16]. Therefore, in terms of later experiments, MLM method is chosen for pre-training models to achieve better understanding from medical text representations. The multilingual BERT is pre-trained by using MLM as well, the introduction will be described in section 2.4.

### 2.3 Domain-Specific pre-trained Language Models

While BERT is proved to efficiently improve the performance compared to traditional NLP methods, it is pre-trained based on the large-scale general domain corpus, such as Wikipedia and BookCorpus [17]. When downstream tasks are situated within a specific domain context, learned representations of BERT might not align with the characteristics of domain-specific text. In such cases, downstream performance often suffers due to a mismatch in vocabulary and syntax.

To address this gap, researchers often apply domain adaptation through a method known as further pre-training or domain-adaptive pre-training (DAPT). In terms of this approach, the language model is further trained on unlabeled data from the target domain using the same self-supervised method, typically masked language modeling (MLM). Domain adaptation via further pre-training has proven effective across a variety of specialized contexts. It provides a flexible solution when labeled data is scarce, as it relies solely on raw domain-specific text to improve the model's capacity to generalize within the target domain, such as finance [16], biomedical [18], law [19], among others. Next, we will discuss the in-domain models related to the medical area.

BioBERT is an English large-scaled pre-trained language model in the biomedical area, which is further pre-trained on bert-base-uncased model over the biomedical corpora from PubMed and PMC [9]. Importantly, it retains the original Wordpiece tokenization used in BERT, which can mitigate the out-of-vocabulary (OOV) issue [20]. Similarly, ClinicalBERT is specialized for English clinical notes of MIMIC-III. This model's performance for predicting readmission of patients is better than BERT. However, this paper also points out that rather than using MIMIC-III ClinicalBERT embeddings, when utilizing ClinicalBERT in hospitals, the model should be retrained to improve its performance on larger EHR datasets [10].

MedRoBERTa.nl [*] is one of the Dutch LLM based on BERT model, which is pre-trained from scratch by using Dutch medical dataset from Amsterdam Medical Centre (AMC) and Medical Centre of the Vrije Universiteit (VuMC) [21]. Through an experiment on domain-specific models and general models, such as BERTje [13] and RoBERTa, in a NER dataset CoNLL-2002 [†], which consists of news articles, it is found that the language in hospital notes is different, and it has a significant effect on the model it produces. Hence, constructing a domain-specific language model highly depends on the pre-training corpus, which indicates that quality, diversity and relevance are crucial.

A BERT-based model for Arabic biomedical NER [22] has been proposed to solve the lack of Arabic biomedical resources utilizing a small-scale biomedical dataset. The model outperformed the NER task better compared to the mBERT and current Ara-BERT models, proving the validity of utilizing small-scale biomedical datasets in the pre-training phase to enhance model performance. Besides, this model was fine-tuned by using a customized silver standard Arabic biomedical corpus, which is tagged with IOB2 encoding format [23], so-called BIO format. In conclusion, this paper shows a detailed procedure of how to pre-train a language model in the absence-resource situation and fine-tune it to realize the NER task.

Thomas et al. [24] presented a BioMed-RoBERTa model to perform the classification of physician notes, screening patients with active bleeding. In this study, they chose the BioMed-RoBERTa model as their language model, which adapts RoBERTa-base to 2.68 million scientific papers from the Semantic Scholar corpus via further pre-training, increasing the proficiency in the biology and medical domain. Patients who did not receive an anticoagulant at the time of admission have been selected from the open source dataset: MIMIC-III, which contains EHR-equivalent patient data. Then, the training set from this dataset was applied to fine-tune the BioMed-RoBERTa model. The results demonstrated that the applicability of alerts increases by 14.8%, while it also reduced the number of patients who would trigger the

---

[*] https://huggingface.co/CLTL/MedRoBERTa.nl
[†] https://www.clips.uantwerpen.be/conll2002/ner/

Multilingual BERT Language Model for Medical Tasks: Evaluation on Domain-Specific Adaptation and Cross-linguality

alerts by 20%. However, the limitation of this study is that they did not use a large training set, which might optimize the classification task.

BioBERTpt[‡] is a deep contextual embedding model for Portuguese to perform the medical NER task, which is developed by pre-trained mBERT on clinical notes and scientific abstracts [25]. In addition to including the real clinical notes from Brazilian hospitals from 2002 to 2018, they integrated the titles and abstracts from Portuguese scientific papers published in Pubmed and Scielo as well. After domain training, they conducted the experiment to find out how domain fine-tuning can affect the downstream task performance, such as NER. In terms of fine-tuning the NER task, a semantically annotated corpus for Portuguese clinical NER, called SemClinBr [26], which includes 1000 labeled clinical data, has been added into the experiment, while another small dataset with IOBIES format, named CLINpt, also has been utilized. The evaluation results show that the BioBERTpt achieves state of the art performance, evidencing that domain-adaptation learning can benefit clinical tasks.

As indicated by Morales-Sánchez et al. [27], the novel early diagnosis approach leveraging unstructured data with LLM yielded a remarkable outcome compared to traditional methods in reducing the missed chances for HIV diagnosis. They trained the roberta model on a Spanish biomedical and clinical corpora. These corpora are derived from several sources, which include publicly available corpora, crawlers and real-world EHR datasets, such as medical crawler [§], a crawler of more than 3,000 URLs belonging to Spanish biomedical and health domains. They discovered that language models outperform traditional NLP methods, which indicates that leveraging effectively unstructured text with language models in the clinical area yields promising results in decreasing missing opportunities of HIV diagnosis. Additionally, they also completed some experiments outside of the research. They chose several Spanish NER datasets and diverse language models, including BioBERT, general-domain Spanish model. The experimental results on these downstream tasks across different chosen models all validate that the Roberta_Bio outperforms other models, which further validates the power of their own model [?].

CancerBERT, a cancer domain-specific language model for extracting breast cancer phenotypes from EHR, is pre-trained based on the BlueBERT, harnessing a corpus from 21,291 breast cancer patients in UMN EHRs [28].

The CancerBERT variants in this paper all outperform other models on the cancer phenotype NER task. Besides, they further investigate the impact of using a customized vocabulary on model performance. Other medical scenarios, such as neuroimaging, radiology, oncology and medical specialty prediction, etc [29, 30, 31, 32], emphasize the value of pre-training phase and the use of domain-specific corpus.

### 2.4 Cross-lingual Transfer

Cross-lingual transfer refers to the ability of a model trained on data from one language (the source language) to perform well on tasks in another language (the target language). This capability is particularly valuable in low-resource settings, where data in the target language is scarce or unavailable. Cross-lingual transfer can be achieved in two main ways. In zero-shot transfer, the model is trained on a task in the source language and directly applied to the target language without additional training [33]. In few-shot transfer, the model is adapted using a small amount of labeled data in the target language.

One of the most widely adopted models for cross-lingual transfer is multilingual BERT (mBERT), introduced by Devlin et al. [4]. It was pre-trained on the concatenated Wikipedia corpora from 104 languages using a shared WordPiece vocabulary. This shared tokenization scheme allows partial lexical overlap between languages, enabling semantically similar tokens to share subword units. Through joint pre-training on multilingual corpora, mBERT encodes similar semantic content into aligned representations across different languages. Furthermore, prior studies [33, 34, 35] have demonstrated the effectiveness of mBERT in transferring linguistic and task-specific knowledge across languages in multilingual settings.

This capability to align representations across multiple languages lays the foundation for analyzing how linguistic factors, such as syntactic distance, affect cross-lingual transfer performance. Suppl.Fig.A.1 illustrates that the language difference is quantified by a strong positive Spearman correlation ($\rho = 0.80$, $p = 8 \times 10^{-40}$) between formal syntactic distance—derived from typological features—and the syntactic representation differences induced by mBERT. This finding indicates that mBERT's internal syntactic representations are sensitive to structural differences between languages, and this strong correlation suggests that syntactic proximity may facilitate more effective cross-lingual transfer [36].

---

[‡]https://huggingface.co/pucpr/biobertpt-all
[§]https://zenodo.org/records/5513237

Multilingual BERT Language Model for Medical Tasks: Evaluation on Domain-Specific Adaptation and Cross-lingualityMultilingual BERT Language Model for Medical Tasks: Evaluation on Domain-Specific Adaptation and Cross-linguality

## 3 Material and Methods

### 3.1 Model and Language Selection

In this study, we chose mBERT as a baseline model for further pre-training for several reasons. First of all, mBERT was pre-trained on a large-scale corpus containing 104 languages, with a vocabulary of approximately 110,000 word units, enabling it to cover multiple languages (Section 2.4). In contrast, BERT was trained only on English corpora, with a vocabulary of only about 30,000 word units, making it difficult to expand to other languages. Therefore, when the downstream tasks involve another language, BERT and its medical variants such as ClinicalBERT and BioBERT struggle to model non-English languages, resulting in a decline in language modeling performance. Furthermore, due to the small-scaled size of our pre-training corpus, it is not feasible to train a monolingual language model from scratch based on BERT—such as the approach taken by MedRoBERTa.nl, which is pre-trained from scratch on a large-scale Dutch clinical corpus.

Besides, although the Dutch medical model MedRoBERTa.nl has been proposed, its monolingual property still limits our exploration of cross-lingual ability. Therefore, we conducted further domain pre-training on a clinical corpus from the Erasmus University Medical Center Rotterdam (Erasmus MC) based on mBERT, which not only evaluated the effectiveness of the corpus in improving Dutch task performance, but also provided stronger scalability and a model baseline for future downstream medical tasks that are transferred to other low-resource languages (such as Romanian or Spanish). Therefore, we chose mBERT as our baseline model, as it has broad language and knowledge coverage and is suitable as a cross-language transfer learning baseline for low-resource medical NLP tasks.

In terms of the language selection, we mainly concentrated on three languages, including Dutch, Romanian and Spanish. First of all, this choice was primarily motivated by the availability of benchmark datasets. Each of selected languages is associated with specific available sub-task datasets for medical NLP: A Dutch HIV classification dataset, a Romanian NER dataset and two Spanish NER datasets. This selection allowed a consistent and realistic experimental setup across multiple languages, enabling competitive cross-lingual comparisons.

The other reason was that there are only a few linguistic resources specific to the biomedical area, especially in Romanian [6]. Additionally, existing pre-trained language models lack domain adaptation to medical text in Romanian, posing more challenges to build Romanian healthcare systems [37]. In addition, Spanish NLP resources are relatively scarce. To better illustrate the situation, on Huggingface Hub, there are 11.35 models and 2.38 datasets per million English speakers, while for Spanish only 2.30 models and 0.59 datasets per million Spanish speakers [7].

The last reason came from the cross-lingual domain adaptation perspective. One of our core hypothesis was that whether a clinical corpus in Dutch can be utilized to improve the performance on downstream tasks in another language that lacks of sufficient clinical resources. In particular, we assessed whether the Erasmus MC clinical dataset, used to further pre-train a Dutch domain-adapted model, can offer cross-lingual benefits when applied to medical tasks in Romanian or Spanish.

### 3.2 Further Pre-training Experiments

To explore the effects of domain adaptation in medical areas, we conducted four further pre-training experiments using masked language modeling (MLM) on mBERT. The corpus selected for further pre-training will be described below. Additionally, details of chosen corpus are illustrated in Table 1.

For Dutch, we trained two separate models to assess the impact of domain-specific pre-training. The first model mBERT-nl-clin used the Erasmus MC clinical corpus, consisting of Dutch hospital records and patient notes, is a model adapted to clinical contexts. The second model mBERT-nl-bio employed a biomedical corpus composed of Dutch medical Wikipedia articles and the EMEA dataset (from European Medicines Agency), representing a more general biomedical domain. This comparison enabled us to examine how in-domain clinical versus biomedical knowledge influences downstream performance.

For Romanian, a low-resource language in the medical domain, we pre-trained mBERT on a biomedical corpus including Romanian EMEA [38], medical Wikipedia content, and resources such as MLPLA and SSLA (collectively referred to as the BioRo corpus[¶]). This experiment explored how further pre-training benefits under-resourced languages with limited annotated data but rich unlabeled biomedical text.

To adapt the general-purpose mBERT to the biomedical and clinical domains, we adopted a domain-adaptive pre-training strategy. This approach, illustrated in Figure 2, inserted an intermediate masked language modeling (MLM) stage between the original pre-training and downstream fine-tuning. Unlike one step pre-training which covered a broad range

---

[¶]https://slp.racai.ro/index.php/resources/bioro-free/

Multilingual BERT Language Model for Medical Tasks: Evaluation on Domain-Specific Adaptation and Cross-linguality

Table 1: Overview of the further pre-training corpora.

| Model | Language | Corpus Source(s) | Tokens |
|---|---|---|---|
| **mBERT-nl-clin** | NL | Erasmus MC clinical corpus | 63.6M |
| **mBERT-nl-bio** | NL | EMEA(NL) | 12.3M |
| | | Wikipedia(NL) | 43.4M |
| **mBERT-ro-bio** | RO | Wikipedia(RO) + BioRo | 9.7M |
| | | EMEA(RO) | 12.0M |
| **mBERT-nl-ro** | NL+RO | Erasmus MC clinical corpus | 63.6M |
| | | Wikipedia(RO) + BioRo | 9.7M |
| | | EMEA (RO) | 12.0M |

of general-domain texts across languages, this stage utilized unlabeled domain-specific corpora to align the model's representations with medical or clinical discourse.

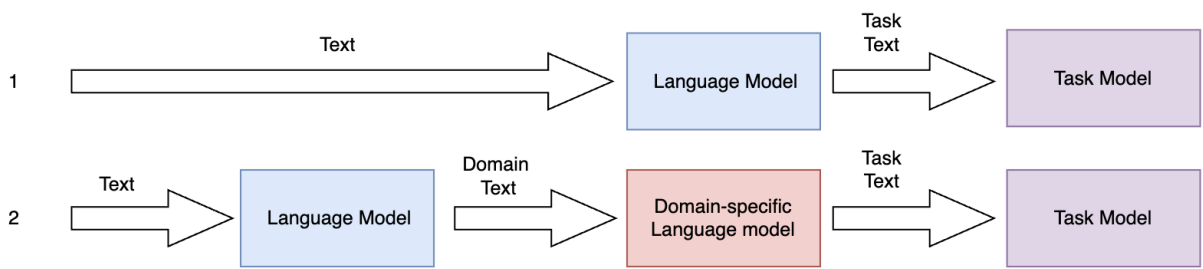

Figure 2: Training schemas: 1. The BERT-based models with one step pre-training on unlabeled text and a fine-tune step with labeled task text. 2. Further pre-training scheme on domain-specific text and a fine-tune step with labeled task text.

### 3.3 Data Preprocessing

The raw corpora used for pre-training were collected from various sources, including Wikipedia, EMEA documents and the clinical notes from Erasmus MC. Before training, non-linguistic elements such as XML tags, HTML labels, and special characters were removed. Duplicate lines and empty lines are also filtered out. Additionally, each line in the corpus represented either a single sentence or a semantically coherent chunk of text.

### 3.4 The training details for pre-training

We employed the masked language model method script to perform domain-adaptive pre-training on the BERT model. All pre-training experiments are conducted on an NVIDIA H100 GPU. The training is conducted using a batch size of 8 per device with gradient accumulation over 4 steps, effectively resulting in a batch size of 32. A learning rate of 2e-5, as recommended in the original BERT study [4], is adopted to ensure stable and effective optimization. In addition, this learning rate setting can efficiently prevent catastrophic forgetting proposed by Sun et al.[39]. The training was run for 5 epochs, with the maximum sequence length of 512 tokens, which corresponds to the input limit of the BERT architecture.

### 3.5 Task-specific Fine-tuning Experiments

To investigate the impact of domain-specific pre-training on downstream medical NLP tasks, we conducted fine-tuning experiments on two representative tasks: HIV classification and Named Entity Recognition (NER).

#### 3.5.1 Sub task 1: HIV classification

First of all, the HIV classification task involved determining whether a patient should be recommended for an HIV test based on their clinical notes, fundamentally a text classification. The dataset was constructed from real-world clinical records, with annotations provided by HIV experts indicating whether each case meets inclusion or exclusion criteria

Multilingual BERT Language Model for Medical Tasks: Evaluation on Domain-Specific Adaptation and Cross-linguality

for HIV tests. The criteria is provided by AWARE.HIV [40] and EuroTest guideline [41]. Models were trained over a balanced dataset with 1838 patient notes with inclusion and exclusion labels.

The experiment followed a stratified 10 folds cross-validation, with the train/test split of 9:1. The experiment process was illustrated in Figure 3. The chosen model details have been described in Table 1. For each model, we calculated the Matthews Correlation Coefficients (MCC) to evaluate the performance of each model in the classification task. The details of MCC can be found in section 3.6.

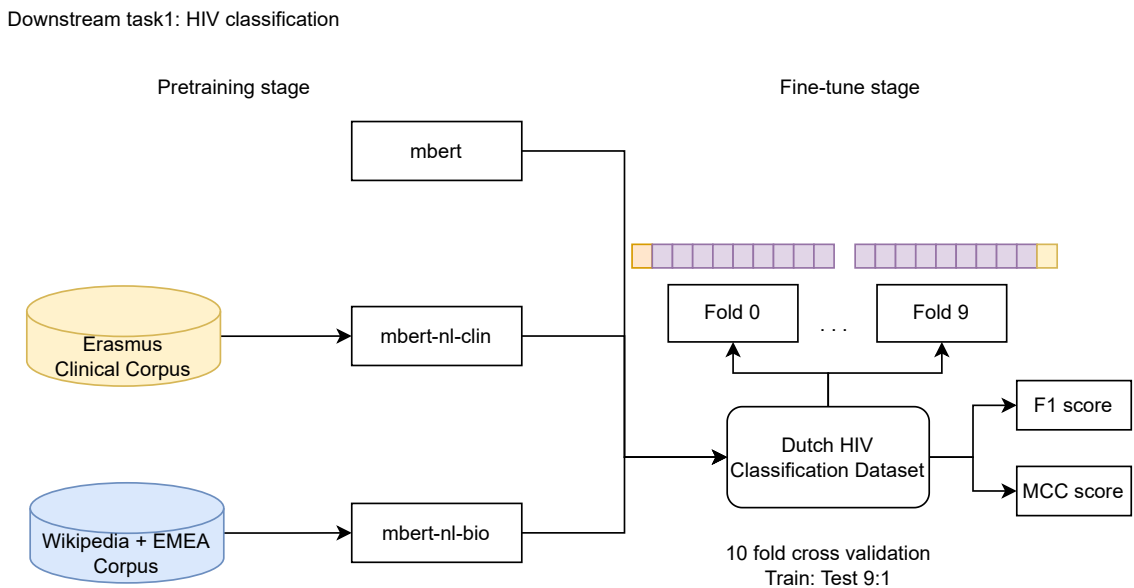

Figure 3: Experiment design of the HIV screening classification task

### 3.5.2 Sub task 2: Named Entity Recognition

Named entity recognition aims to identify and classify medically relevant entities in unstructured, such as diseases, medications, and anatomical terms. In this study, we focused on Romanian and Spanish medical texts, where annotated entities follow a standard BIO tagging scheme. This task allowed the model to learn how to categorize structured information from unstructured medical narratives, and serves as a meaningful benchmark to assess the effect of domain adaptation and cross-lingual ability in a low-resource setting.

**Romanian NER task** The Romanian MoNERo dataset is composed of three entity types: Chemicals and Drugs (CHEM), Disorders (DISO), Anatomy (ANAT) and Procedures (PROC). The description of each entity has been clarified by Maria et al.[42]. It comprises 497 sentences, each containing one or more annotated entities across various categories. In total, the RoNER dataset contains 1,175 annotated named entities, including 522 CHEM, 395 DISO, 192 ANAT, and 66 PROC entities.

- CHEM: amino acid, peptide, protein, antibiotic, biologically active substance, chemical, clinical drug, hormone, organic chemical, pharmacologic substance, receptor, steroid, vitamin;
- DISO: acquired abnormality, anatomical abnormality, cell or molecular dysfunction, congenital abnormality, disease or syndrome, experimental model of disease, finding, injury or poisoning, sign or symptom;
- ANAT: body location or region, body part, organ, or organ component, body substance, body system, cell, fully formed anatomical structure, tissue
- PROC: diagnostic procedure, health care activity, laboratory procedure, molecular biology research technique, therapeutic or preventive procedure.

The experiment for the Romanian NER also adopted 10-fold cross-validation. Fine-tuning was performed for 15 epochs with a batch size of 16, a learning rate of 2e-5, as recommended by Sun et al. [39] for preventing catastrophic forgetting.

Multilingual BERT Language Model for Medical Tasks: Evaluation on Domain-Specific Adaptation and Cross-linguality

Since NER datasets often suffer from label imbalance, with a disproportionately high number of 'O-' tags compared to 'B-' and 'I-' tags , the F1-score is a more appropriate evaluation metric than precision or recall alone.

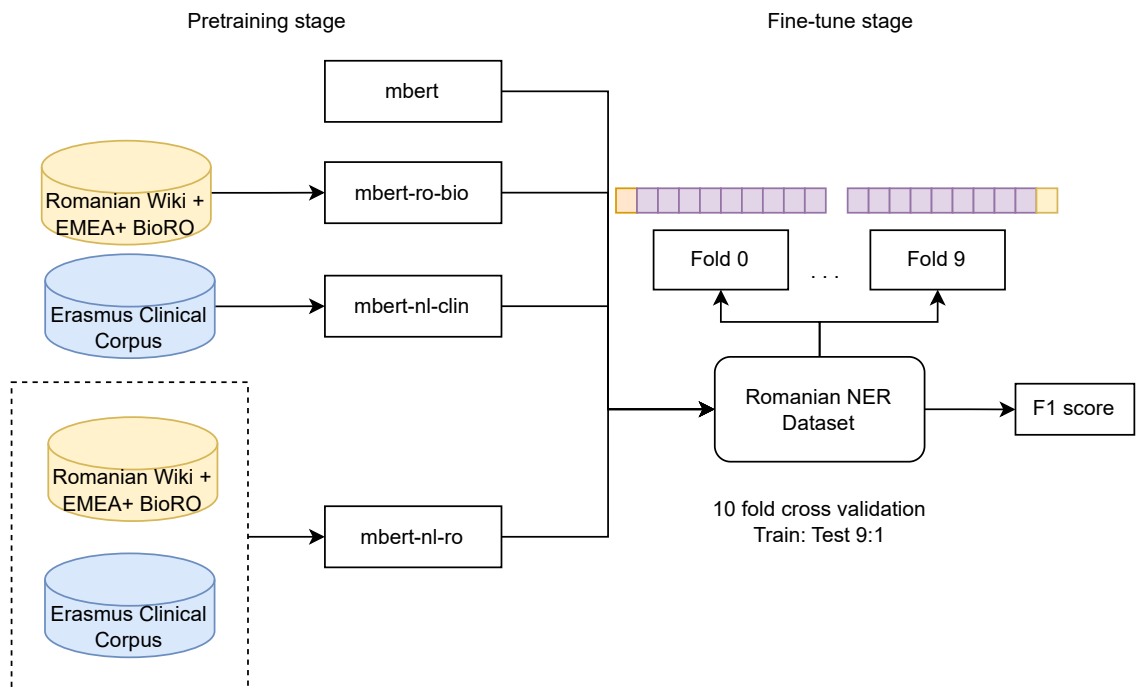

Figure 4: Experiment design of the Romanian NER task

### 3.5.3 Spanish NER task

Firstly, the **CANTEMIST-NER** dataset [||] is part of a shared task focused on Named Entity Recognition (NER) of tumor morphology. It is collected from 1301 oncological clinical case reports written in Spanish, with tumor morphology mentions manually annotated and mapped by clinical experts to a controlled terminology [43]. In this dataset, only entities related to morphological tumor concepts were annotated as MORFOLOGIA_NEOPLASIA. Totally, this dataset is composed of 19397 sentences, and there are 6347 entities labeled as MORFOLOGIA_NEOPLASIA.

Another Spanish NER dataset we utilized is **PharmaCoNER**, which is composed of manually classified collection of clinical case studies derived from the Spanish Clinical Case Corpus. The annotation of the entire set of entity mentions was carried out by domain experts. The annotation process includes the following 4 entity types: NORMALIZABLES, NO_NORMALIZABLES, PROTEINAS and UNCLEAR[44]. This dataset comprises 8129 sentences. The number of total named entities is 3786, including 2282 NORMALIZABLES, 1392 PROTEINAS, 88 UNCLEAR, 24 NO_NORMALIZABLES.

- NORMALIZABLES: The chemical mentions can be manually normalized to a certain concept identifier.
- NO_NORMALIZABLES: The chemical mentions cannot be manaully normalized to a certain identifier.
- PROTEINAS: mentions of proteins/genes, including peptides, peptide hormones & antibodies.
- UNCLEAR: cases involving clinically relevant mentions of general substance categories, including certain drug formulations, general treatments, chemotherapy regimens, and vaccines.

For Spanish NER sub task, the fine-tune parameter settings are the same as Romanian NER task. As illustrated in Figure 5, the bsc-bio-ehr-es model is a biomedical-clinical language model for Spanish, which is an open source model

---
[||]https://huggingface.co/datasets/PlanTL-GOB-ES/cantemist-ner

Multilingual BERT Language Model for Medical Tasks: Evaluation on Domain-Specific Adaptation and Cross-linguality

in Huggingface. This model was further pre-trained on Spanish biomedical and clinical corpus, created by Carrino et al.[?]. To evaluate the model performance on this sub task, we also chose average F1-score from 10 folds.

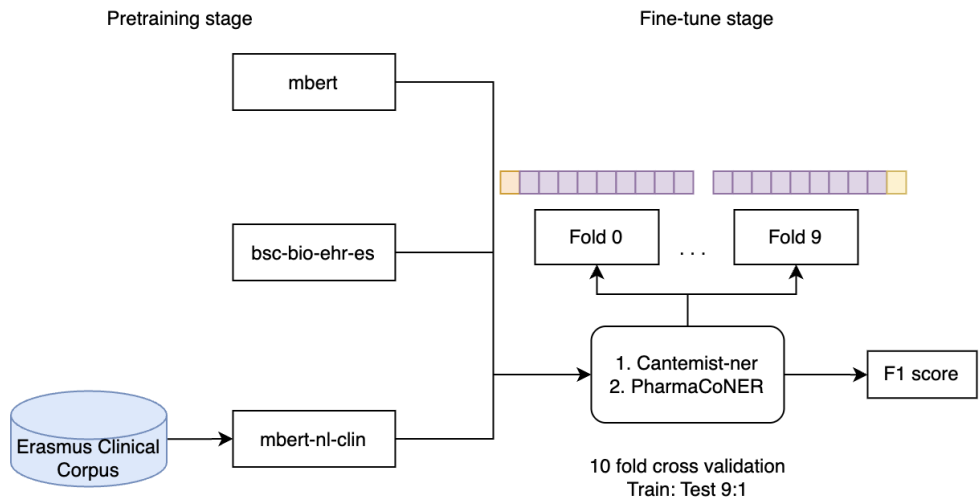

Figure 5: Experiment design of the Spanish NER task

### 3.6 Evaluation Metrics

For HIV screening classification task, we utilized Matthews Correlation Coefficient (MCC). It takes into account all four categories of the confusion matrix: true positives, true negatives, false positives, and false negatives. It produces a value between -1 and 1, where 1 indicates perfect prediction, 0 indicates random guessing, and -1 indicates total disagreement between predictions and actual labels. MCC is particularly useful for evaluating performance on imbalanced classification tasks [45]. The mathematic equation of MCC is explained below. For NER classification task, we used macro-F1 score to evaluate the model performance. The score is computed globally over all instances and all class labels [46], which can better reflect the overall ability of models.

$$MCC = \frac{TP \times TN - FP \times FN}{\sqrt{(TP+FP)(TP+FN)(TN+FP)(TN+FN)}} \qquad (4)$$

In addition, in order to rigorously assess whether the performance differences between two models are statistically significant, the Wilcoxon sign-ranked test was applied based on the selected metrics per fold between models. Wilcoxon sign-ranked test is a non-parametric statistical test that evaluates whether two paired samples come from distributions with same median. In addition, it does not assume the normality of distribution, and it can reduce the influence of outliers. A p-value below 0.05 is considered statistically significant, indicating that the performance difference between models is unlikely due to random deviation.

## 4 Results

### 4.1 Sub task 1: Dutch HIV classification

In order to evaluate how domain adaptation affects the performance of mBERT on medical NLP tasks, we selected three models to fine-tune using the labeled dataset. In addition, we calculate the mean score of each model's MCC and macro-F1 output, which indicates the mBERT-nl-clin reaches the best average performance, while mBERT obtains the worst average performance.

First of all, both mBERT-nl-bio and mBERT-nl-clin showed a better MCC score compared to mBERT, with the p-values 0.027 and 0.010 respectively, highlighting the benefit of medical domain adaptation. Notably, mBERT-nl-clin outperforms mBERT-nl-bio with p-value of 0.002, suggesting that general biomedical model was not as useful as the clinical model when it came to a clinical downstream task with many clinical texts. This implies that within a broad

Multilingual BERT Language Model for Medical Tasks: Evaluation on Domain-Specific Adaptation and Cross-linguality

specific domain (e.g. medical), differences in subdomains (general biomedical versus clinical) can still significantly affect the model performance in downstream tasks.

In addition, from a clinical perspective, experts emphasized the importance of recall, also referred to as sensitivity. The reason can be explained by, in real-world scenarios, failing to identify a patient at risk of HIV (predicting Exclusion when the correct label is Inclusion) could have serious consequences. As shown in Table 2, mBERT-nl-clin obtained the best recall score, while mBERT obtained the lowest recall score.

| Model | MCC | F1 | Recall |
|---|---|---|---|
| mBERT | 0.412 | 0.697 | 60.19 |
| mBERT-nl-bio | 0.437 | 0.708 | 64.53 |
| mBERT-nl-clin | **0.469** | **0.726** | **65.67** |

Table 2: The evaluation results of Dutch HIV classification

### 4.2 Sub task 2: Named Entity Recognition

#### 4.2.1 Romanian NER

Table 3 illustrates the average macro-F1 score across 10-fold cross-validation for each model. To assess whether the observed performance differences between models are statistically significant, we conducted Wilcoxon signed-rank tests across the 10-fold results.

| Model | F1 | Recall | Precision |
|---|---|---|---|
| mBERT | 73.51 | 75.42 | 71.12 |
| mBERT-nl-clin | 71.80 | 73.56 | 70.08 |
| mBERT-nl-ro | 75.62 | 78.11 | 73.34 |
| mBERT-ro-bio | **76.80** | **78.52** | **75.17** |

Table 3: The evaluation results of the Romanian NER task

mBERT-ro-bio significantly outperformed mBERT, with p-values of 0.004, indicating that domain-adapted pre-training on Romanian biomedical data provides consistent performance gains. mBERT-nl-ro also significantly outperformed mBERT (0.020), highlighting the benefit of bilingual domain pre-training. However, the comparison between mBERT-ro-bio and mBERT-nl-ro (p = 0.375) revealed no statistically significant difference, suggesting their performance does not lead to an inferior effect, despite their different pre-training corpora. Interestingly, the mBERT-nl-clin versus mBERT comparison (p = 0.322) was not significant, confirming that clinical data in Dutch alone does not improve performance on Romanian biomedical tasks. Therefore, the result cannot validate the cross-lingual ability between Dutch and Romanian.

Furthermore, in order to gain deeper insights of model performance across different entity types, we conduct a fine-grained analysis of F1 score per category. As shown in Table 4, both in-domain models mBERT-ro-bio and mBERT-nl-ro performed superior than the other two out-of-domain models. To be specific, mBERT-ro-bio reached the best performance on CHEM and DISO entities, while mBERT-nl-ro obtained the best performance on ANAT and PROC.

| Model | ANAT | CHEM | DISO | PROC |
|---|---|---|---|---|
| mBERT | 73.64 | 79.80 | 70.61 | 14.47 |
| mBERT-nl-clin | 69.97 | 77.87 | 69.94 | 3.83 |
| mBERT-ro-bio | 74.44 | **82.90** | **75.63** | 19.74 |
| mBERT-nl-ro | **77.74** | 81.16 | 72.84 | **25.41** |

Table 4: F1 scores of different models across entity types in the Romanian NER task

Multilingual BERT Language Model for Medical Tasks: Evaluation on Domain-Specific Adaptation and Cross-linguality

### 4.2.2 Spanish NER

In terms of Spanish NER task, we fine-tuned models over two datasets: Cantemist-ner and PharmaCoNER. The results will be discussed respectively in the following paragraphs. In the end, the similarities and differences of results between these two Spanish NER datasets were described.

**Cantemist-ner** Table 5 shows the average F1 scores across 10 folds on Cantemist-ner dataset. The bsc-bio-ehr-es model obtained the best performance in this experiment. Firstly, results in Table 5 showed that the performance of bsc-bio-ehr-es is better than mBERT with p-value 0.002. Additionally, bsc-bio-ehr-es outperformed another two models on Recall and Precision. The Dutch model mBERT-nl-clin can perform better in Cantemist-ner compared to mBERT with p-value 0.023, which indicated the potential of cross-lingual ability from Dutch to Spanish. Since Cantemist-ner dataset only included one entity type, the model performance on this entity type reflected the overall ability. Moreover, the mBERT-nl-clin model gained better results on all the evaluation metrics compared to mBERT.

| Model | F1 | Recall | Precision |
|---|---|---|---|
| mBERT | 82.25 | 83.62 | 80.92 |
| mBERT-nl-clin | 82.63 | 83.86 | 81.44 |
| bsc-bio-ehr-es | **84.07** | **85.91** | **82.30** |

Table 5: The evaluation results of the Spanish NER task on Cantemist-ner dataset

**PharmaCoNER** Table 6 displays the average F1 scores of each model. We observed that bsc-bio-ehr-es significantly outperformed mBERT and mBERT-nl-clin with p < 0.005 across comparisons. These conformed that domain-adaptation yielded improved model performance on PharmaCoNER dataset. Besides, mBERT-nl-clin showed a better result than mBERT, with p-value of 0.002.

| Model | F1 | Recall | Precision |
|---|---|---|---|
| mBERT | 86.76 | 87.84 | 86.78 |
| mBERT-nl-clin | 87.70 | 88.21 | 86.80 |
| bsc-bio-ehr-es | **89.60** | **90.93** | **88.02** |

Table 6: The evaluation results of the Spanish NER task on PharmaCoNER dataset

Across both Cantemist-ner and PharmaCoNER datasets, the results consistently highlighted the importance of domain-adaptation. In each case, models pre-trained on in-domain corpora in Spanish settings outperformed general-purpose baselines like mBERT.

Additionally, we observed that mBERT-nl-clin performed better than mBERT in Catemist-ner (p-value = 0.023) and in PharmaCoNER (p-value = 0.002), suggesting that Dutch pre-training can offer additional benefits in low-resource Spanish settings, potentially due to cross-lingual transfer effects between related domains.

Similarly, we evaluated the performance of each model across diverse entity types in PharmaCoNER, to gain more insights from the results. As illustrated in Table 7, the in-domain model outperformed the other two models across all entities. Especially for the entity NO_NORMALIZABLES, in-domain model significantly outperforms mBERT and mBERT-nl-clin, with the gap of 18.28% and 21.79%. In addition, we can find that mBERT-nl-clin works slightly better than bsc-bio-ehr-es in terms of the UNCLEAR entity type. The reasoning will be discussed in Section 5.

| Model | NORMALIZABLES | NO_NORMALIZABLES | PROTEINAS | UNCLEAR |
|---|---|---|---|---|
| mBERT | 90.27 | 5.80 | 84.02 | 74.04 |
| mBERT-nl-clin | 91.19 | 2.29 | 84.96 | **82.20** |
| bsc-bio-ehr-es | **93.35** | **24.08** | **86.79** | 81.84 |

Table 7: F1 scores of different models across entity types in PharmaCoNER dataset

Multilingual BERT Language Model for Medical Tasks: Evaluation on Domain-Specific Adaptation and Cross-linguality

## 5 Discussion

### 5.1 Further Analysis of HIV classification

To further understand the output of models, we conducted a qualitative case study using Local Interpretable Model-agnostic Explanations (LIME). Since mBERT and mBERT-nl-bio obtain lower recall scores compared to mBERT-nl-clin, we further examine the interpretability of model predictions to understand which textual features each model relies on. We specifically select a case where mBERT and mBERT-nl-bio fail to include the patient with HIV risk, while mBERT-nl-clin makes the correct inclusion decision.

The instance is labeled as Inclusion, indicating that the patient is at risk of HIV and should be included for further HIV testing. Interestingly, mBERT predicts exclusion with a confidence score of 0.378 wrongly, while mBERT-nl-clin correctly predicts inclusion with a high confidence score of 0.776. The LIME was applied to observe the most influential tokens from the text for both models [47, 48]. We visualized the top 10 tokens to observe which tokens the model emphasizes.

As demonstrated in Figure 6, "HBV", "lues", "Behandelgeschiedenis" are the most influential tokens contributing positively to the inclusion prediction by mBERT-nl-clin. These terms are highly relevant in the HIV case: "HBV" refers to the hepatitis B virus, "lues" is the historical term for syphilis, and "Behandelgeschiedenis" means treatment history. To be more specific, "HBV" and "lues" are both important HIV indicators listed in the EuroTest guideline. On the contrary, the less meaningful tokens received lower confidence scores.

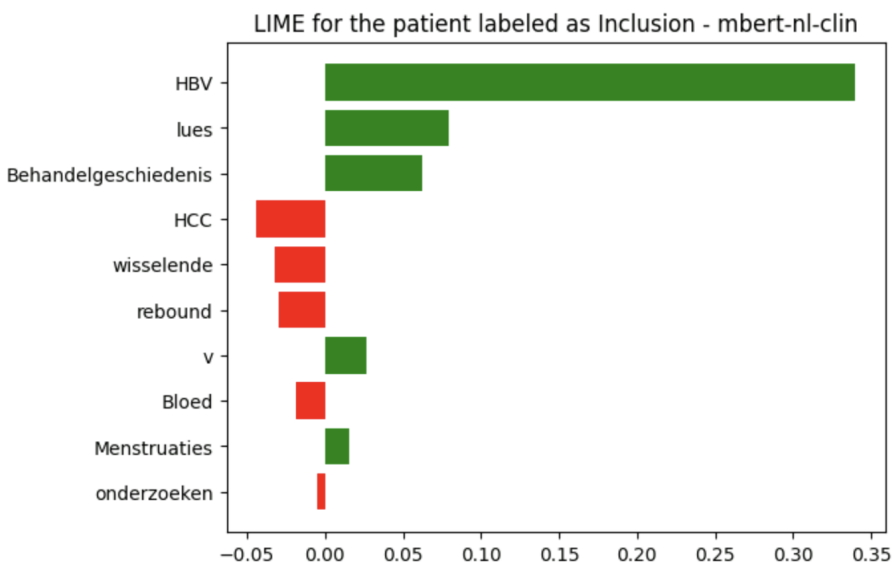

Figure 6: The LIME analysis of mBERT-nl-clin on the HIV screening classification task. The green bars indicate Inclusion prediction, and the red bars indicate Exclusion prediction. The x-axis is the confidence score.

In contrast, mBERT and mBERT-nl-bio assigned positive weights to tenofovir for the exclusion class, despite it being a key treatment drug associated with Inclusion. Moreover, mBERT assigned relevance to multiple terms unrelated to HIV inclusion criteria, such as "DNA", "gepland" (planned), "Indien" (if), yet incorrectly treated them as evidence supporting the inclusion decision. This inconsistency indicates that the model lacks clinical knowledge in the decision-making process. Although mBERT-nl-bio predicted the instance mistakenly, it correctly assigned weight to "Lamivudine", which is a treatment drug for HIV and Hepatitis-B. Here, "M" from the clinical notes shows no meaning, while "fl" is an abbreviation for follicular lymphoma.

The LIME technique allowed us to visualize the contributions of each token during predictions, offering which parts of the input the model attends to most. Through this error analysis, we can tell that the domain-adaptation model like mBERT-nl-clin was equipped to capture clinical features that align with the HIV indicators list set by HIV experts. By contrast, the baseline mBERT tended to focus on superficial or irrelevant textual cues, leading to lower performance.

Multilingual BERT Language Model for Medical Tasks: Evaluation on Domain-Specific Adaptation and Cross-linguality

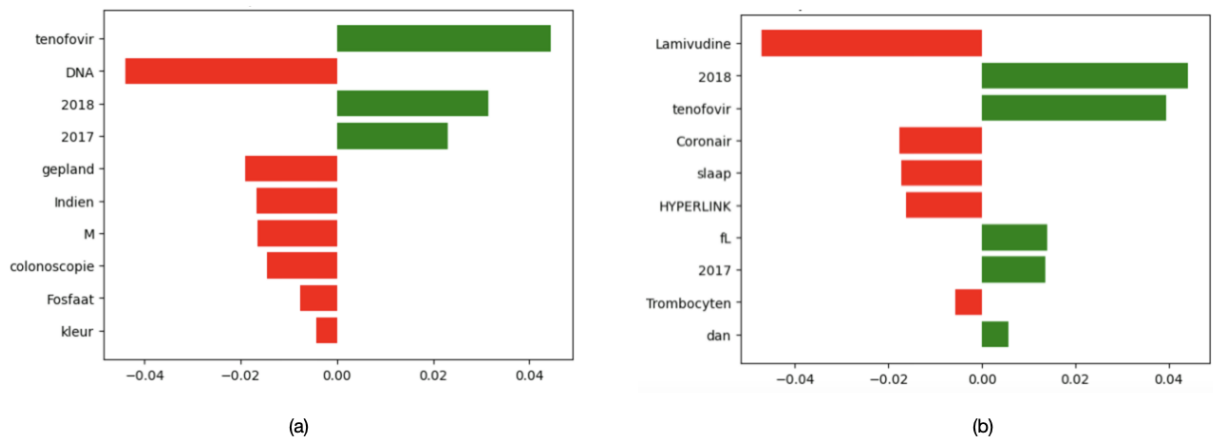

Figure 7: The LIME analysis of mBERT(a) and mBERT-nl-bio(b) on the HIV screening classification task. The green bars indicate Inclusion prediction, and the red bars indicate Exclusion prediction. The x-axis is the confidence score.

### 5.2 Fine-Grained Entity-Level Analysis For NER

In the results section, we provided a preliminary analysis of the model's performance across different entity types. In this section, we conducted a fine-grained investigation to better understand models' performance across diverse entity types.

For the Romanian NER task, we can see in-domain models mBERT-ro-bio and mBERT-nl-ro outperformed across all the entity categories. However, all models consistently received relatively low F1 scores on the PROC entity type. Upon further investigation, we hypothesized several possible reasons for the consistently bad performance on PROC. The first point was that PROC entities are underrepresented compared to other entity types such as CHEM and DISO, with the latter occurring almost six times more frequently in the dataset. Besides, we assumed that PROC entities exhibit higher contextual variability. This diversity made it difficult for the model to consistently learn a coherent representation.

In order to prove our assumptions, we calculated the intra-entity similarity by using extracted embeddings from the final hidden layer of mBERT-ro-bio. In particular, we computed the average cosine similarity between all entity vectors within the same category. We observed that PROC reached a lower intra-entity similarity compared to other entities, indicating its contextual embeddings are more dispersed in the semantic space. The results can be found in the Suppl.Tab.A.1. Moreover, we visualized the embeddings of each entity type by applying t-distributed Stochastic Neighbor Embedding (t-SNE) as shown in Figure 8. t-SNE is a dimensionality reduction technique that projects high-dimensional data into low-dimensional space while preserving the local structure, which is commonly used to visualize the embeddings produced by language models [49]. The PROC entity appeared highly dispersed and intermixed with other entity types, forming no distinct cluster. By combining the intra-entity similarity and T-SNE visualization, the result suggests that the model struggles to learn a consistent representation of PROC type, due to their structural complexity and contextual variability.

For PharmaCoNER dataset, the worst performance was consistently presented on the NO_NORMALIZABLES entity type across all models. From a data distribution perspective, this can be attributed to severe class imbalance. For example, while the PROTEINS entity type contains 2,963 instances, NO_NORMALIZABLES comprises only 50, making it challenging for the model to learn meaningful representations for this minority class. Furthermore, as shown in Figure 9, NO_NORMALIZABLES entities failed to form a clearly defined cluster in the embedding space. By contrast, other entity types exhibited a clear spatial separation. The scattered distribution of NO_NORMALIZABLES was likely caused by the high contextual variability and semantic inconsistency, which might make it more difficult to be recognized accurately. Combined with the class imbalance of NER datasets, these findings reinforce the explanation for why they perform poorly on a specific class and highlight the challenge of modeling these entity types effectively.

### 5.3 Impact of Domain Adaptation

For NER tasks across two different languages, we extracted contextual embeddings generated by in-domain models and mBERT, then compute intra-entity similarity using cosine similarity. We did not consider the minority entity types, such as PROC in Romanian, since they are not representative enough. A higher intra-entity similarity indicates that entities of the same category can be represented more densely in the embedding space. The results show that the representations generated by in-domain models exhibit consistently higher similarity than those generated by mBERT. This finding suggests that domain-adapted models can generate more coherent and semantically similar representations for medical entities. Among all the experiments across different languages, the domain-adaptation shows promising power for medical downstream tasks as shown in Table 8.

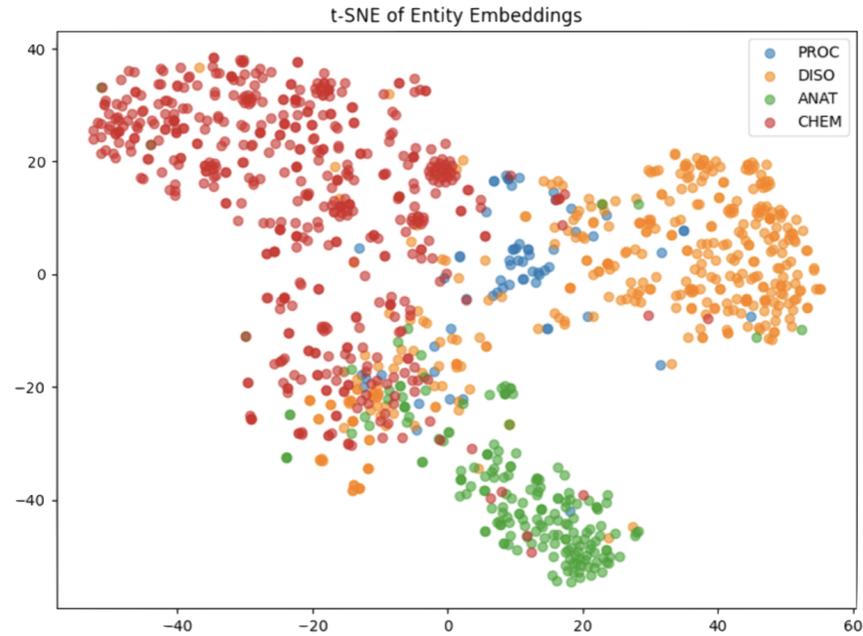

Figure 8: t-SNE projection of entity embeddings from mBERT-ro-bio for the Romanian NER task.

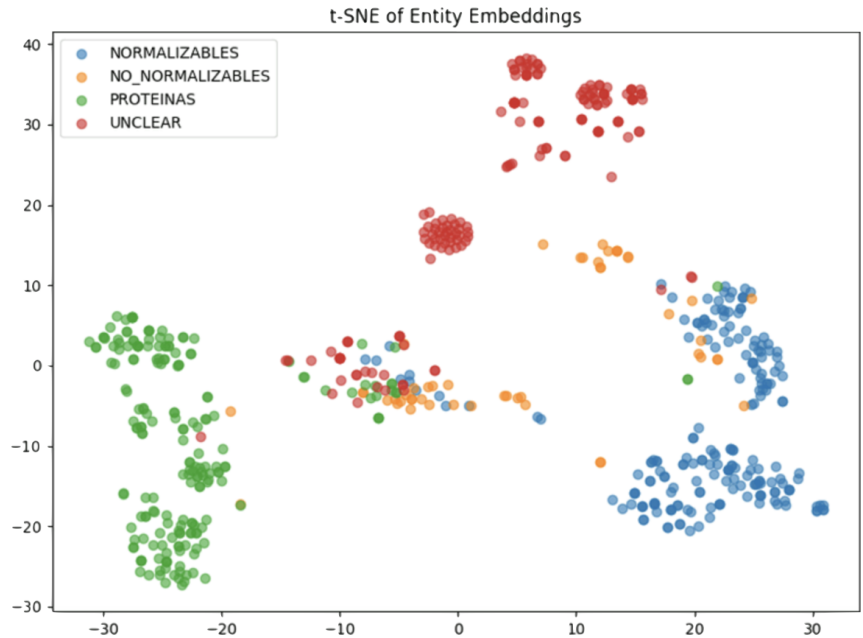

Figure 9: t-SNE projection of entity embeddings from mBERT on the Spanish PharmaCoNER dataset.

Multilingual BERT Language Model for Medical Tasks: Evaluation on Domain-Specific Adaptation and Cross-linguality

Table 8: Performance comparison between domain-adapted models and the baseline mBERT model

| Task | Domain-adapted Model | Metric | Improvement vs mBERT | p-value |
|---|---|---|---|---|
| Dutch HIV Classification | mBERT-nl-bio | MCC | +0.025 | 0.027 |
|  | mBERT-nl-clin | MCC | +0.057 | 0.002 |
| Romanian NER | mBERT-nl-ro | F1 | +0.024 | 0.020 |
|  | mBERT-ro-bio |  | +0.036 | 0.004 |
| Spanish Cantemist-NER | bsc-bio-ehr-es |  | +0.019 | 0.002 |
| Spanish PharmaCoNER | bsc-bio-ehr-es |  | +0.021 | 0.002 |

### 5.4 Impact of Further pre-training on Clinical versus Biomedical Corpora

We created two diverse models based on different corpus, so called mBERT-nl-bio and mBERT-nl-clin. The mBERT-nl-bio was further pre-trained on bio-medical corpus, including Wikipedia medical articles, and drug instructions from the EMEA dataset. The mBERT-nl-clin was further pre-trained on real-world clinical notes from Erasmus MC. In section 3.5.1, both models were fine-tuned over the Dutch HIV classification dataset, to classify whether the patients should be excluded or included for a HIV test recommendation. The results show that mBERT-nl-clin outperforms mBERT-nl-bio (MCC gain 0.031, p-value 0.010),, indicating that within the same language and broader medical domain, differences among sub-domains can have a measurable impact on the downstream tasks performance. This also highlights that if the domain of downstream tasks can align with the domain adaptation in pre-training stage, an optimal performance can be reached.

### 5.5 Cross-Lingual Transferability

For the Spanish NER task, the results in Table 9 show promising potential for transferring Dutch medical knowledge to Spanish. The results on datasets Cantemist-NER and PharmaCoNER show that mBERT-nl-clin obtained better performance than mBERT. This effect is particularly notable on the Cantemist-NER dataset, where mBERT-nl-clin showed statistically significant improvement, suggesting that pre-training on Dutch clinical data can enhance performance even in a different target language when the domain is shared. One possible reason is the syntactic difference in mBERT between Dutch and Spanish is smaller than that of Dutch to Romanian, as shown in Suppl.Fig. A.1. In short, better cross-lingual transfer ability appears to be associated with the syntactic similarity between source and target languages. Based on the current experiments we have completed, medical knowledge in Dutch can be better transferred to Spanish than to Romanian.

Table 9: F1 score improvement of mBERT-nl-clin over the baseline mBERT model regarding cross-lingual transferability

| Task | Model | F1 improvement | p-value |
|---|---|---|---|
| Romanian NER | mBERT-nl-clin | -1.4% | 0.322 |
| Spanish Cantemist-NER | mBERT-nl-clin | +0.4% | **0.023*** |
| Spanish PharmaCoNER | mBERT-nl-clin | +0.94% | **0.002*** |

## 6 Conclusions

In this study, we aimed to investigate the potential of domain adaptation and cross-lingual ability of mBERT language model. We created models by applying the further pre-training method across multiple languages and diverse domains, and then fine-tuned these models on the labeled datasets over different downstream tasks, e.g. binary HIV classification and NER task.

The domain-adaptation power was validated across three languages and two different tasks. Among all experiments across different languages, the domain adaptation shows promising power for medical downstream tasks. First of all, the mBERT-nl-bio and mBERT-nl-clin after further pre-training on the medical domain corpus outperformed mBERT, which was only trained on a large general domain corpus. Similarly, in terms of NER tasks in Romanian and Spanish, the domain-adapted models, both mBERT-ro-bio and bsc-bio-ehr-es obtained better performances compared to the baseline model mBERT. We can conclude that further pre-training on a domain-specific corpus shows a significant potential to enhance downstream task performance. Further, we observed that the model further pre-trained on clinical corpora outperformed the one on general biomedical corpora when the downstream task is specifically clinical.

The experiments for cross-lingual ability were conducted across two languages: Spanish and Romanian. We propose that cross-lingual transferability mainly relies on the syntactic similarity between the source and target languages. The closer syntactic similarity between two languages, the better the transferability can be achieved. Our experimental results demonstrate that Dutch shows stronger transfer performance to Spanish, while the transfer from Dutch to Romanian is weaker.

Future work includes expanding the experiments to other low-resourced languages to evaluate the generalizability of our findings across different linguistic families, such as Polish and Hungarian. Additionally, the mBERT-nl-clin model we have created can be fine-tuned on different clinical downstream tasks, e.g. summarizing radiology and pathology reports. Besides, due to privacy concerns, only the Dutch clinical corpus was accessible for pre-training, whereas the Romanian and Spanish corpora were limited to open-source general medical texts. In the future, access to clinical corpora in other languages could enable a more detailed investigation into the sub-domain effects. Furthermore, we will continue to improve the generative AI models' cross-lingual ability according to the findings in this study. The code repository is https://github.com/ErasmusMC-Bioinformatics/AI4HIV-YinghaoLuo.

# 7 Acknowledgement

This study was funded by the ErasSupport grant provided by Erasmus University Medical Center Rotterdam. The #aware.hiv project was supported by the Dutch Federation Medical Specialist (SKMS) (grant number: 59825822) and by an unrestricted investigator-initiated study grant from Gilead Sciences and ViiV Healthcare. The industry was not involved in the study design, data collection, analysis, interpretation, or submission for publication. We gratefully acknowledge Viola Woeckel for her guidance on data selection and transfer procedures, and for safeguarding all aspects related to data privacy and security.

Multilingual BERT Language Model for Medical Tasks: Evaluation on Domain-Specific Adaptation and Cross-linguality

## A  Supplementary materials

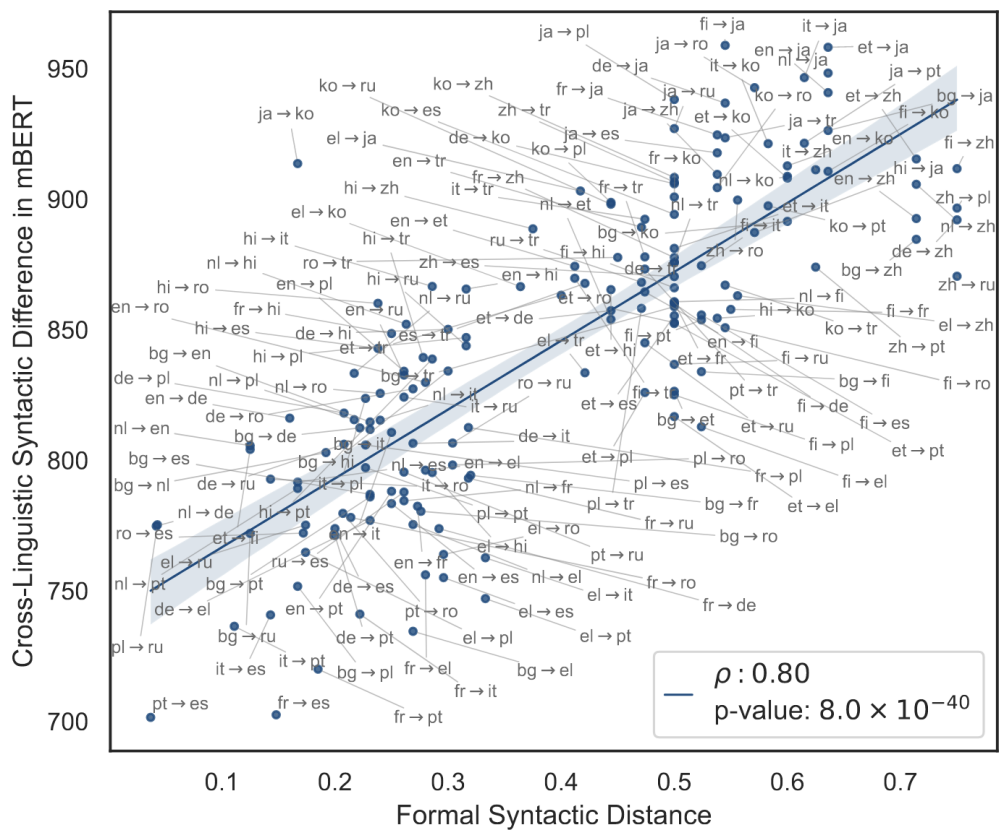

Supplementary Figure A.1: Correlation between formal syntactic distance and syntactic representation differences in mBERT across language pairs (adapted from Pires et al. [50]). The lower the value on the y-axis, the more similar two languages are. In other words, languages that are syntactically closer (e.g. German and Dutch) are encoded more similarly by mBERT, while syntactically distant pairs (e.g. Dutch and Chinese) exhibit larger representational divergence.

| Model | DISO | ANAT | CHEM |
|---|---|---|---|
| mBERT | 0.4623 | 0.4934 | 0.5018 |
| mBERT-ro-bio | 0.6034 | 0.6004 | 0.5219 |

Supplementary Table A.1: Intra-entity similarity of entity representations of Romanian NER

| Model | PROTEINAS | NORMALIZABLES | UNCLEAR |
|---|---|---|---|
| mBERT | 0.7827 | 0.6044 | 0.6785 |
| bsc-bio-ehr-es | 0.7377 | 0.8524 | 0.8314 |

Supplementary Table A.2: Intra-entity similarity of entity representations(PharmaCoNER)